\documentclass{llncs}
\usepackage{makeidx}  % allows for indexgeneration
\pdfoutput=1
\usepackage{graphicx}
\usepackage[lofdepth,lotdepth]{subfig}
\usepackage{algorithm}
\usepackage{algpseudocode}
\usepackage{amsmath, amssymb}

\graphicspath{{img//}}

\newcommand{\argmax}{\operatornamewithlimits{argmax}}

\begin{document}
\mainmatter              % start of the contributions
\title{Bootstrapping Monte Carlo Tree Search with an Imperfect Heuristic}
\titlerunning{UCT-Aux}  % abbreviated title (for running head)
\author{Truong-Huy Dinh Nguyen \and Wee-Sun Lee \and Tze-Yun Leong}
\authorrunning{Nguyen et al.}   % abbreviated author list (for running head)
%
%%%% modified list of authors for the TOC (add the affiliations)
\tocauthor{Truong-Huy D. Nguyen, Wee-Sun Lee, Tze-Yun Leong (National University of Singapore)}
\institute{National University of Singapore, Singapore 117417 \\
\email{trhuy, leews, leongty@comp.nus.edu.sg}}

\maketitle              % typeset the title of the contribution

\begin{abstract}
We consider the problem of using a heuristic policy to improve the value approximation by the Upper Confidence Bound applied in Trees (UCT) algorithm in non-adversarial settings such as planning with large-state space Markov Decision Processes. Current improvements to UCT focus on either changing the action selection formula at the internal nodes or the rollout policy at the leaf nodes of the search tree. In this work, we propose to add an auxiliary arm to each of the internal nodes, and always use the heuristic policy to roll out simulations at the auxiliary arms. The method aims to get fast convergence to optimal values at states where the heuristic policy is optimal, while retaining similar approximation as the original UCT in other states. We show that bootstrapping with the proposed method in the new algorithm, UCT-Aux, performs better compared to the original UCT algorithm and its variants in two benchmark experiment settings. We also examine conditions under which UCT-Aux works well.
\end{abstract}

\section{Introduction}
\label{sec:intro}

Monte Carlo Tree Search (MCTS)~\cite{coulom2006efficient}, or more specifically Upper Confidence Bound applied in Trees (UCT)~\cite{kocsis2006}, is a state-of-the-art approach to solving large state-space planning problems. Example applications of the UCT algorithm in the games domain include Go~\cite{coulom2006efficient,ICML07_silver,chaslot2010adding}, General Game Playing~\cite{2008finnsson}, Real-Time Strategy Game~\cite{balla2009uct}, etc.

The algorithm estimates the value of a state by building a search tree using simulated episodes, or \textit{rollouts}, via interactions with the simulator. Instead of sampling every branch equally, the goal is to focus samplings in tree branches that are more promising. In particular, UCT achieves that by choosing the action, or \textit{arm} if its parent node is regarded as a multi-armed bandit, with the highest estimated upper bound to simulate at every internal node, and randomly selects actions after leaving the tree to finish the rollout.

Because UCT uses random sampling to discover nodes with good return, it could take a long time to achieve good performance. To address this problem, many enhancements have been used to improve the search control of the algorithm by either (1) tweaking the action selection formula at the internal nodes~\cite{bouzy2004monte,ICML07_silver,2008finnsson,chaslot2010adding,coquelin2007}, or/and (2) designing better-informed rollout policies in place of random sampling at the leaf nodes~\cite{ICML07_silver,chaslot2010adding}.

We consider the problem of using a heuristic function to improve the approximated value function computed by UCT. Taking the approaches above, the first method is to initialize new tree nodes with heuristic values and the second is to use the chosen heuristic to roll out simulations at the leaf nodes. As intended, these two methods could greatly influence the search control by guiding it into more promising regions that are determined by the heuristic. However, when the heuristic function does not accurately reflect the prospect of the states, it could feed the algorithm with false information, thereby leading the search into regions that should be kept unexplored otherwise. 

In this work, we propose a novel yet simple enhancement method. Given a heuristic in the form of an imperfect policy $ \pi $, the method adds an additional arm at every internal node of the search tree. This special arm is labeled by the action suggested by $ \pi $ and once selected, rolls out the rest of the sampling episode using $ \pi $. If the policy $\pi$ works well at a state, we expect it to quickly give a good estimate of the value of the state without relying too much on the other arms. % that are further expanded in the tree search.
The method aims to get fast convergence to optimal values at states where the heuristic policy is optimal, while retaining similar approximation as the original UCT in other states.

We compared this method with two aforementioned techniques in two domains, namely Obstructed Sailing, an extension of the original Sailing problem previously used to measure UCT's performance in~\cite{kocsis2006}, and Sheep Savior, a large state-space MDP that characterizes a generic two-player collaborative puzzle game. The results showed that UCT-Aux the new algorithm (\textbf{Aux} for \textit{auxiliary} arms) significantly outperforms its competitors when coupled with reasonable heuristics.

One nice property of this method is that it does not affect the implementation of other bootstrapping techniques: No modification of the action selection formula nor the rollout policy at any leaf nodes except for the added arms is required. This allows different sources of bootstrapping knowledge to be combined into one algorithm for more performance boost.

The rest of the paper is structured as follows. We first give a brief overview of MDP, UCT and its popular enhancements before presenting UCT-Aux. Next, we describe two experimental setups for comparing the agents' performance and analyze the results. We also identify the common properties of the heuristics used in two experimental domains and provide some insights on why UCT-Aux works well in those cases. Finally, we conclude the paper by discussing the possible usage of UCT-Aux.

\section{Background}
\label{sec:backgrounds}

\subsection{Markov Decision Process}
\label{sec:mdp}

A Markov Decision Process characterizes a planning problem with tuple $ (S, A, T, R) $, in which
\begin{itemize}
\item $ S $ is the set of all states,
\item $ A $ is the set of all available actions,
\item $ T_{a}(s, s') = P(s_{t+1}=s' \vert s_{t}=s, a_{t}=a) $ is the probability that action $ a \in A $ in state $ s \in S $ at time $ t $ will lead to state $s' \in S$ at time $ t+1 $. 
\item $ R_{a}(s,s')$ is the immediate reward received after the state transition from $s$ to $s'$ triggered by action $a$.
\end{itemize}

An action \textit{policy} $ \pi $ is a function, possibly stochastic, that returns an action $ \pi(s) $ for every state $ s \in S $. In infinite-horizon discounted MDPs, the objective is to choose an action policy $ \pi^* $ that maximizes some cumulative function of the received rewards, typically the expected discounted sum $ \sum_{t=0}^{\infty} \gamma^t R_{a^*}(s_t, s_{t+1}) $ with $0 \leq \gamma < 1$ being the discount factor. An MDP can be effectively solved using different methods, one of which is the value iteration algorithm based on the Bellman's equation~\cite{Bel}. The algorithm maintains a value function $V(s)$, where $s$ is a state, and iteratively updates the value function using the equation
\[ V_{t+1}(s)=\max_{a}\left(\sum_{s'}T_{a}(s,s')(R_{a}(s,s')+\gamma V_{t}(s'))\right). \]
This value iteration algorithm is guaranteed to converge to the optimal value function $V^*(s)$, which gives the optimal expected cumulative reward of running the optimal policy from state $s$.

The optimal value function $V^{*}$ can be used to construct the optimal policy by taking action $a^{*}$ in state $s$ such that $a^{*}=\mbox{argmax}_{a} \left\lbrace \sum_{s'} T_{a}(s,s')V^{*}(s') \right\rbrace$. The optimal $Q$-function is constructed from $V^{*}$ as follows:
\[ Q^{*}(s,a) = \sum_{s'} T_{a}(s,s')(R_{a}(s,s') + \gamma V^{*}(s')).\]
$Q^{*}(s,a)$ denotes the maximum expected long-term reward of an action $a$ when executed in state $s$.% instead of just telling how valuable a state is, as does $V^{*}$.

One key issue that hinders MDPs and Value Iteration from being widely used in real-life planning tasks is the large state space size (usually exponential in the number of state variables) that is often required to model realistic problems.

\subsection{Upper Confidence Bound Applied to Trees (UCT)}
\label{sec:mcts}

UCT~\cite{kocsis2006} is an anytime algorithm that approximates the state-action value in real time using Monte Carlo simulations. It was inspired by Sparse Sampling~\cite{kearns2002}, the first near-optimal policy whose runtime does not depend on the size of the state space. The approach is particularly suitable for solving planning problems with very large or possibly infinite state spaces.

\begin{figure}[h]
\centering
\includegraphics[scale=0.4]{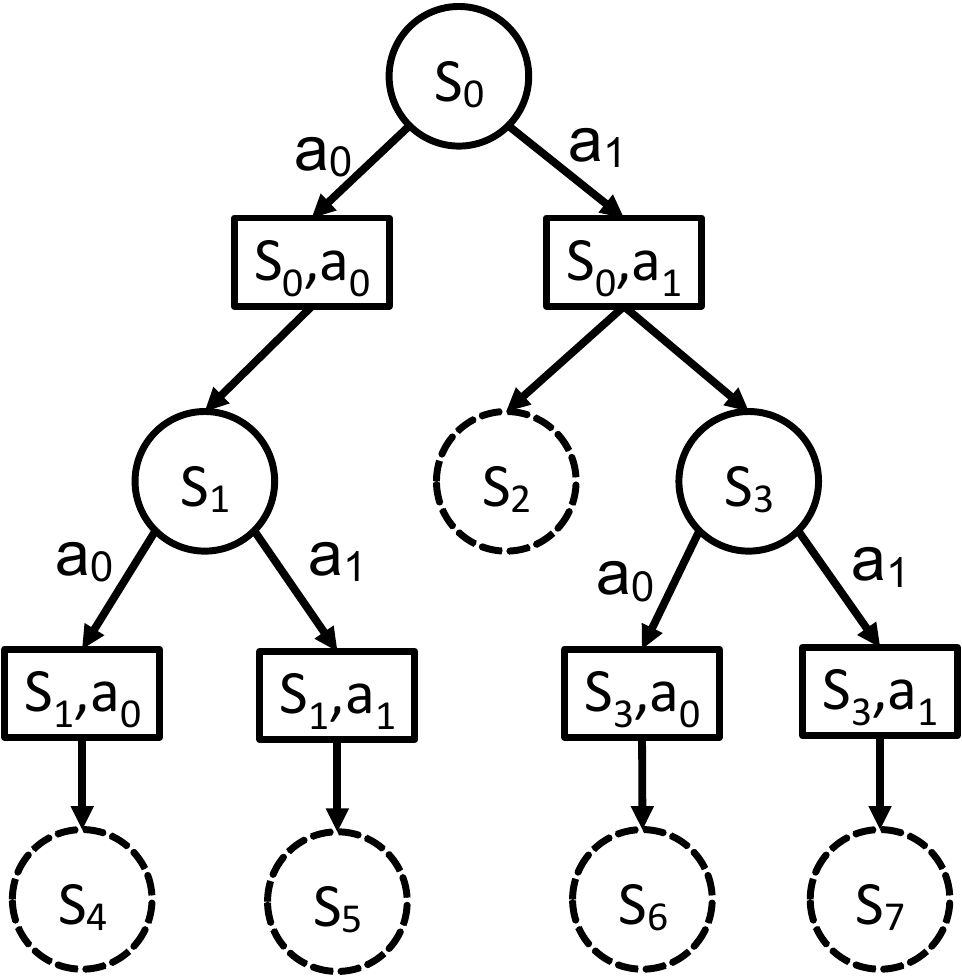} 
\caption{A sample UCT search tree with two valid actions $ a_0 $ and $ a_1 $ at any state. Circles are \textit{state} nodes and rectangles are \textit{state-action} nodes; solid state nodes are \textit{internal} while dotted are \textit{leafs}.}
\label{fig:mcts}
\end{figure}

The algorithm searches forward from a given starting state, building up a tree whose nodes alternate between reachable future states and state-action pairs (Figure~\ref{fig:mcts}). State nodes are called \textit{internal} if their child state-action pairs have been expanded and \textit{leaf} otherwise. Starting with a root state, the algorithm iteratively rolls out simulations from this root node; each time an internal node is encountered, it is regarded as a multi-armed bandit and UCB1~\cite{auer02:finite} is used to determine the action or \textit{arm} to sample, i.e., the edge to traverse. In particular, at an internal node $ s $, the algorithm selects an action according to
\begin{equation}
\label{eq:ucb-uct}
\pi_{UCT}(s) = \argmax_{a}  \left\lbrace  Q_{UCT}(s,a) + 2C_{p}\sqrt{\frac{\log n(s)}{n(s,a)}} \right\rbrace,
\end{equation}
in which
\begin{itemize}
\item $Q_{UCT}(s,a) $ is the estimated value of state-action pair $ (s,a) $, taken to be the weighted average of its children's values.
\item $ C_p >0 $ is a suitable hand-picked constant.
\item $ n(s) $ is the total number of rollouts starting from $ s $.
\item $ n(s,a) $ is the number of rollouts that execute $ a $ at $ s $.
\end{itemize} 

At the chosen child state-action node, the simulator is randomly sampled for a next state with accompanying reward; new states automatically become leaf nodes. From the leaf nodes, rollouts are continued using random sampling until a termination condition is satisfied, such as reaching terminal states or simulation length limit. Once finished, the returned reward propagates up the tree, with the value at each parent node being the weighted average of its child nodes' values; suppose the rollout executes action $ a $ at state $ s $ and accumulates reward $ R(s,a)  $ in the end.
\begin{itemize}
\item at state-action nodes, $ n(s,i) = n(s,i)+1 $ and $ Q_{UCT}(s,a)= Q_{UCT}(s,a) + \frac{1}{n(s,a)} (R(s,a) - Q_{UCT}(s,a))$
\item at state nodes, $ n(s) = n(s)+1 $.
\end{itemize}
Typically one leaf node is converted to internal per rollout, upon which its child state-action nodes are generated. When the algorithm is terminated, the root's arm with highest $ Q_{UCT}(s,a) $ is returned\footnote{In practice, returning the arm with highest $ n(s,a) $ is also a common choice.}.

When used for infinite-horizon discounted MDPs, the search can be cut off at an $ \epsilon_0 $-horizon. Given any $ \epsilon >0 $, with $ \epsilon_0$ small enough, the algorithm is proven to converge to the arm whose value is within the $ \epsilon $-vicinity of the optimal arm~\cite{kocsis2006}.

\subsection{Enhancement methods}
\label{sec:gelly-algos}

In vanilla UCT, new state-action nodes are initialized with uninformed default values and random sampling is used to finish the rollout when leaving the tree. Given a source of prior knowledge, Gelly and Silver~\cite{ICML07_silver} proposed two directions to bootstrap UCT: 
\begin{enumerate}
\item Initialize new action-state nodes with $ n(s,a) = n_{prior}(s,a) $ and $ Q_{UCT}(s,a) = Q_{prior}(s,a) $, and
\item Replace random sampling by better-informed exploration guided by $ \pi_{prior} $. 
\end{enumerate}

We refer to these two algorithms as UCT-I (UCT with new nodes \textbf{i}nitialized to heuristic values) and UCT-S (UCT with \textbf{s}imulations guided by $ \pi_{prior}$); UCT-IS is the combination of both methods. UCT-I and UCT-S can be further tuned using domain knowledge to mitigate the flaw of a bad heuristic and amplify the influence of a good one by adjusting the dependence of the search control on the heuristic at internal nodes. In this work, we do not investigate the effect of such tuning to ensure a fair comparison between techniques when employed as is.

In the same publication~\cite{ICML07_silver}, the authors proposed another bootstrapping technique, namely Rapid Action Value Estimation (RAVE), which we do not examine in this work. The technique is specifically designed for domains in which an action from a state $ s $ has similar effect regardless of when it is executed, either at $ s $ or after many moves. RAVE uses the All-Moves-As-First (AMAF) heuristic~\cite{bruegmann1993monte} instead of $ Q_{UCT}(s,a) $ in Equation~\ref{eq:ucb-uct} to select actions. Many board games such as Go or Breakthrough~\cite{2008finnsson} have this desired property. In our experiment domains, RAVE is not applicable, because the actions are mostly directional movements, e.g., $ \{N,E,S,W\} $, thus tied closely to the state they are performed at.

\section{UCT-Aux: Algorithm}
\label{sec:algorithm}

Given an added policy $ \pi $, we propose a new algorithm UCT-Aux that follows the same search control as UCT except for two differences.

\begin{figure}[h]
\centering
\includegraphics[scale=0.4]{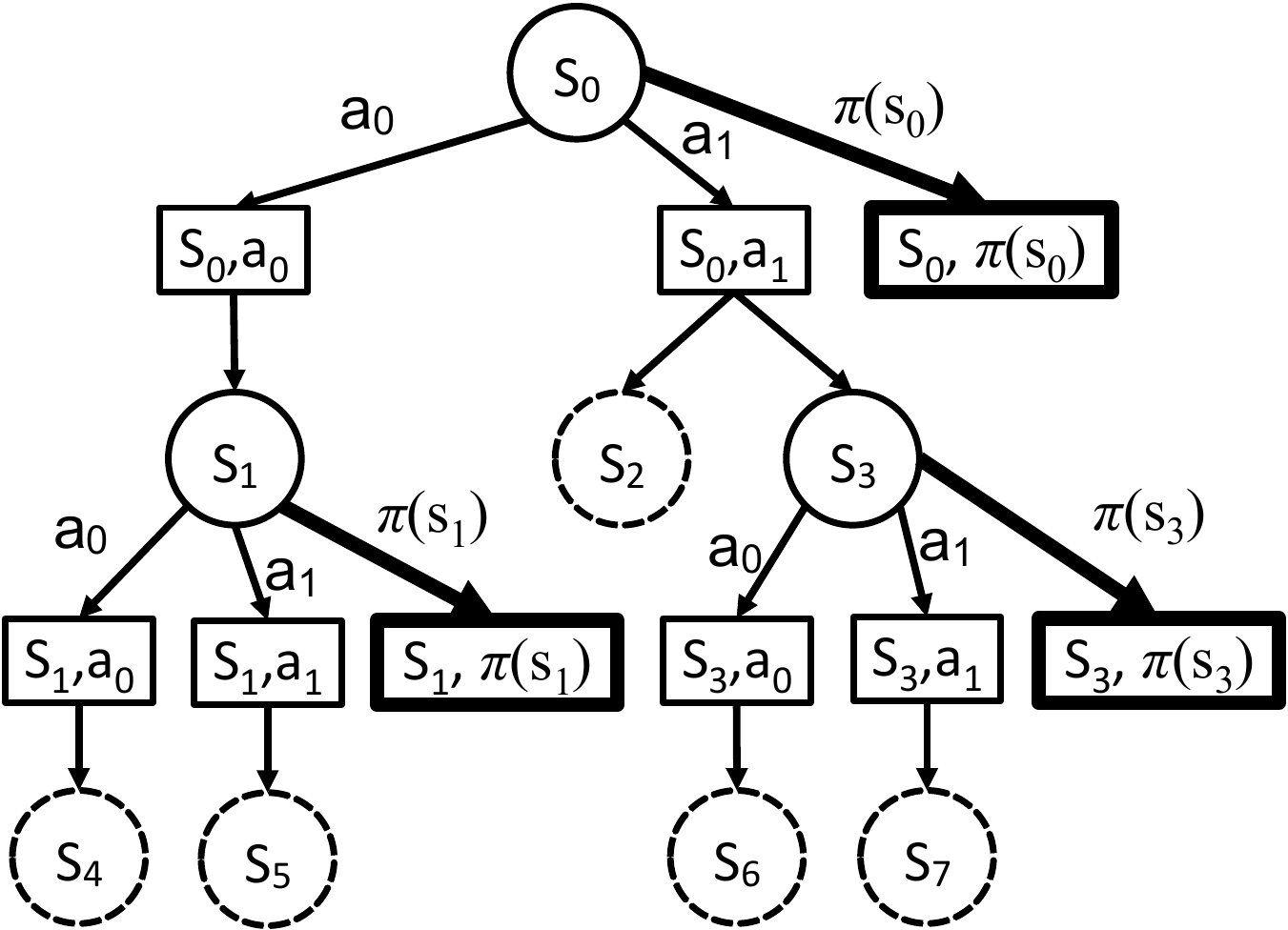} 
\caption{Sample search tree of UCT-Aux.}
\label{fig:uctplus-tree}
\end{figure}

\begin{enumerate}
\item At every internal node $ s $, besides $ |A(s)| $ normal arms with $ A(s) $ being the set of valid actions at state $ s $, an additional arm labeled by the action $ \pi(s) $ is created (Figure~\ref{fig:uctplus-tree}).
\item When this arm is selected by Equation~\ref{eq:ucb-uct}, it stops expanding the branch but rolls out a simulation using $ \pi $; value update is carried out from the auxiliary arm up to the root as per normal.
\end{enumerate}

The method aims to better manage mixed-quality heuristics. If the heuristic $\pi$'s value estimation at a state is good, we expect the added arm to dominate the distribution of rollouts and quickly give a good estimate of the state's value without the need to inspect other arms. Otherwise, the search control will focus rollouts in ordinary arms, thus retaining similar approximation as vanilla UCT.

For stochastic heuristic policies, at every internal node, not one but $ \kappa $ auxiliary arms are added, with $ \kappa $ being the number of actions $ a_{\pi} $ such that $ P(\pi(s)=a_{\pi}) > 0 $. As such, the number of arms at internal nodes is bounded by $ 2|A| $ since $ \kappa \leq |A| $.

\subsection*{Convergence Analysis}
\label{sec:convergence}

We will show that regardless of the added policy's quality, UCT-Aux converges in finite-horizon MDPs\footnote{As mentioned in~\cite{kocsis2006}, for use with discounted infinite-horizon MDPs, the search tree can be cut off at the effective $ \epsilon_0 $-horizon with $ \epsilon_0 $ being the desired accuracy at root.}. The proof follows closely that of UCT analysis by treating the auxiliary arms as any other ordinary arms. As a recap, UCT convergence analysis revolves around the analysis of non-stationary multi-armed bandits with reward sequences satisfying some drift conditions, which is proven to be the case for UCT's internal nodes with appropriate choice of bias sequence $ C_p $\footnote{Empirically, $ C_p $ is often chosen to be an upper bound of the accumulated reward starting from the current state.}. In particular, the drift conditions imposed on the payoff sequences go as follows:
\begin{itemize}
\item The expected values of the averages $ \overline{X}_{in}=\frac{1}{n}\sum^n_{t=1} X_{it}  $ must converge for all arms $ i $ with $ n $ being the number of pulls and $ X_{it} $ the payoff of pull $ t $. Let $ \mu_{in}=E[\overline{X}_{in}] $ and $ \mu_i = lim_{n\rightarrow \infty}\mu_{in} $.
\item $ C_p >0 $ can be chosen such that the tail inequalities $ P(\overline{X}_{i,n(i)} \geq \mu_i + c_{t,n(i)}) \leq t^{-4} $ and $ P(\overline{X}_{i,n(i)} \leq \mu_i - c_{t,n(i)}) \leq t^{-4} $ are satisfied for $ c_{t,n(i)} = 2C_p \sqrt{\frac{\ln t}{n(i)}} $ with $ n(i) $ being the number of times arm $ i $ is pulled up to time $ t $.
\end{itemize}

Firstly, we will show that all internal nodes of UCT-Aux have arms yielding rewards satisfying the drift conditions. Suppose the horizon of the MDP is $ D $, the number of actions per state is $ K $ and the heuristic policy is deterministic ($ \kappa = 1 $); this can be proven using induction on $ D $. Note that i.i.d. payoff sequences satisfy the drift conditions trivially due to Hoeffding's inequality.
\begin{itemize}
\item $ D=1 $: Suppose the root has already been expanded, i.e., become internal. It has $ K+1 $ arms, which either lead to leaf nodes (ordinary arms) or return i.i.d. payoffs (auxiliary arm). Since leaf nodes have i.i.d. payoffs, all arms satisfy drift conditions. 
\item $ D>1 $: Assume that all internal state nodes under the root have arms satisfying the drift conditions, e.g., $ s_1 $ and $ s_3 $ in Figure~\ref{fig:uctplus-tree}. Consider any ordinary arm of the root node (the added arm's payoff sequence is already i.i.d.), for instance, $ (s_0,a_1) $.
Its payoff average is the weighted sum of payoff sequences in all leafs and state-action nodes on the next two levels of the subtree, i.e., leaf $ s_2 $, arms $ (s_3,a_0),(s_3,a_1) $ and $ (s_3,\pi(s_3)) $, all of which satisfy drift conditions due to either the inductive hypothesis or producing i.i.d. payoffs. Theorem 4 in~\cite{kocsis2006} posits that the weighted sum of payoff sequences conforming to drift conditions also satisfies drift conditions; therefore, all arms originating from the root node satisfy drift conditions.%\hfill $ \square $ 
\end{itemize}

As a result, the theorems on non-stationary bandits in~\cite{kocsis2006} hold for UCT-Aux's internal nodes as well. Therefore, we can obtain similar results to Theorem 6 of~\cite{kocsis2006}, with the difference being statistical measures related to the auxiliary arms such as $ \mu_{aux} $ and $ \Delta_{aux} $, i.e., the new algorithm's probability of selecting a suboptimal arm converges to zero as the number of rollouts tends to infinity. 

\section{Experiments}
\label{sec:experiments}

We compare the performance of UCT-Aux against UCT, UCT-I, UCT-S and UCT-IS in two domains: Obstructed Sailing and Sheep Savior. Obstructed Sailing extends the benchmark Sailing domain by placing random blockage in the map; the task is to quickly move a boat from one point to a destination on a map, disturbed by changing wind, while avoiding obstacles. Sheep Savior features a two-player maze game in which the players need to herd a sheep into its pen while protecting it from being killed by two ghosts in the same environment. 

\section{Obstructed Sailing}
\label{sec:sailing}

\begin{figure}[h]
\centering
\subfloat[][Obstructed Sailing sample map]{
\includegraphics[height=180pt]{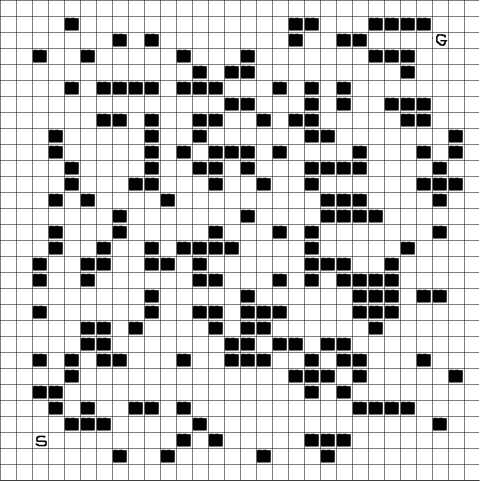} 
\label{fig:sample-sailing}
}
\qquad
\subfloat[][SailTowardsGoal]
{
\includegraphics[height=180pt]{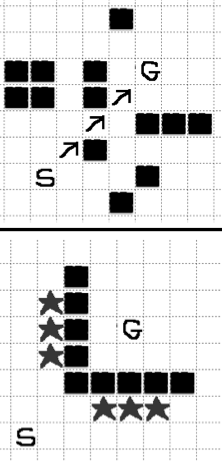} 
\label{fig:bad-heu}
}
\caption{Obstructed Sailing domain; (a) a randomized starting configuration, (b) SailTowardsGoal heuristic produces near-optimal estimates/policies in good cases but misleads the search control in others.}
\label{fig:sailing}
\end{figure}

The Sailing domain, originally used to evaluate the performance of UCT~\cite{kocsis2006}, features a control problem in which the planner is tasked to move a boat from a starting point to a destination under certain disturbing wind conditions. In our version, there are several obstacles placed randomly in the map (see Figure~\ref{fig:sample-sailing}).

In this domain, the state is characterized by tuple $ \langle x,y, b, w_{prev}, w_{curr} \rangle $ with $ (x,y) $ being the current boat position, $ b $ the  current boat posture or direction, $ w_{prev} $ the previous wind direction and $ w_{curr} $ the current wind direction. Directions take values in \textit{\{N, NE, E, SE, S, SW, W, NW\}}, i.e. clockwise starting from North. The controller's valid action set includes all but the directions against $ w_{curr} $, out of the map or into an obstacle. After each time step, the wind has roughly equal probability to remain unchanged, switch to its left or its right~\cite{sailing1996}.

Depending on the relative angle between the action taken and $ w_{curr} $, a cost from 1 to 4 minutes is incurred. Additionally, changing from a port to a starboard tack or vice versa causes a tack delay of 3 minutes. In total, an action can cost anywhere from 1 to 7 minutes, i.e., $ C_{min} = 1 $ and $ C_{max} = 7 $~\cite{sailing1996}. We model the problem as an infinite-horizon discounted MDP with discount factor $ \gamma=0.99 $.

\subsection{Choice of heuristic policies}

A simple heuristic for this domain is to select a valid action that is closest to the direction towards goal position regardless of the cost, thereafter referred to as SailTowardsGoal. For instance, in the top subfigure of Figure~\ref{fig:bad-heu}, at the starting state marked by ``S", if current wind is not SW, SailTowardsGoal will move the boat in the NE direction; otherwise, it will execute either N or E.

This heuristic is used in UCT-I and UCT-IS by initializing new state-action nodes with values
\[ n_{STG}(s,a) \leftarrow 1 \]
\[ Q_{STG}(s,a) \leftarrow C(s,a) + C_{min} \frac{1 - \gamma^{d(s',g)+1}}{1 - \gamma} \]
with $ C(s,a) $ being the cost of executing action $ a $ at state $ s $ and $ d(s',g) $ the minimum distance between next state $ s' $ and goal position $ g $. The initialized value can be seen as the minimum cost incurred when all future wind directions are favorable for desired movement. For UCT-S, the random rollouts are replaced by $ \pi(s) = \argmax_a Q_{STG}(s,a) $.

\textbf{Heuristic quality.} This heuristic works particularly well for empty spaces, producing near-optimal plans if there are no obstacles. However, it could be counterproductive when encountering obstacles. In the bottom subfigure of Figure~\ref{fig:bad-heu}, if a rollout from the starting position is guided by SailTowardsGoal, it could be stuck oscillating among the starred tiles, thus giving inaccurate estimation of the optimal cost.

\subsection{Setup and results}

\begin{figure}[h]
\centering
\includegraphics[width=250pt]{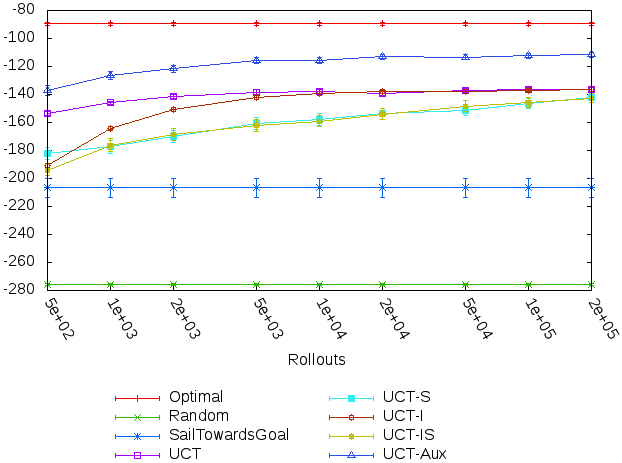} 
\caption{Performance comparison of UCT, UCT-S, UCT-I, UCT-IS and UCT-Aux when coupled with the heuristic SailTowardsGoal; y-axis is the reward average with error bars being the standard errors of the means.}
\label{fig:sailing-result}
\end{figure}

The trial map size is 30 by 30, with fixed starting and goal positions at respectively $ (2,2) $ and $ (27,27) $ (Figure~\ref{fig:sample-sailing}). We generated 100 randomized instances of the map, where obstacles are shuffled by giving each grid tile $ p=0.4 $ chance to be blocked~\footnote{We tried with different values of $ p \in \{0.05, 0.1, 0.2, 0.3, 0.5\} $ and they all yield similar results as Figure~\ref{fig:sailing-result}; the detailed charts are not presented due to space constraint.}. Each instance is tried five times, each of which with different starting boat postures and wind directions. 

All UCT variants (UCT, UCT-I, UCT-S, UCT-IS and UCT-Aux) use the same $ C_p=C_{max} / (1 - \gamma) = 700 $ and the search horizon\footnote{The search horizon is chosen to be long enough so that the cost accumulated after the horizon has small effect to the total cost.} is set to be 300; an optimal path should not be very far from 60 steps as most actions move the boat closer to the goal. The exact optimal policy is obtained using Value Iteration. Note that the performance of Optimal agent varies because of the randomization of starting states (initial boat and wind direction) and map configurations.

Given the same number of samplings, UCT-Aux outperforms all competing UCT variants, despite the mixed quality of the added policy SailTowardsGoal when dealing with obstacles (Figure~\ref{fig:sailing-result}). Note that without parameter tuning, both UCT-I and UCT-S are inferior to vanilla UCT, but between UCT-I and UCT-S, UCT-I shows faster performance improvement when the number of samplings increases. The reason is because when SailTowardsGoal produces inaccurate heuristic values, UCT-I only suffers at early stage while UCT-S endures misleading guidance until the search reaches states where the policy yields more accurate heuristic values. The heuristic's impact is stronger in UCT-S than UCT-I: UCT-IS's behavior is closer to UCT-S than UCT-I.

\section{Sheep Savior}
\label{sec:sheep-savior}

\begin{figure}[h]
\centering
\includegraphics[scale=0.51]{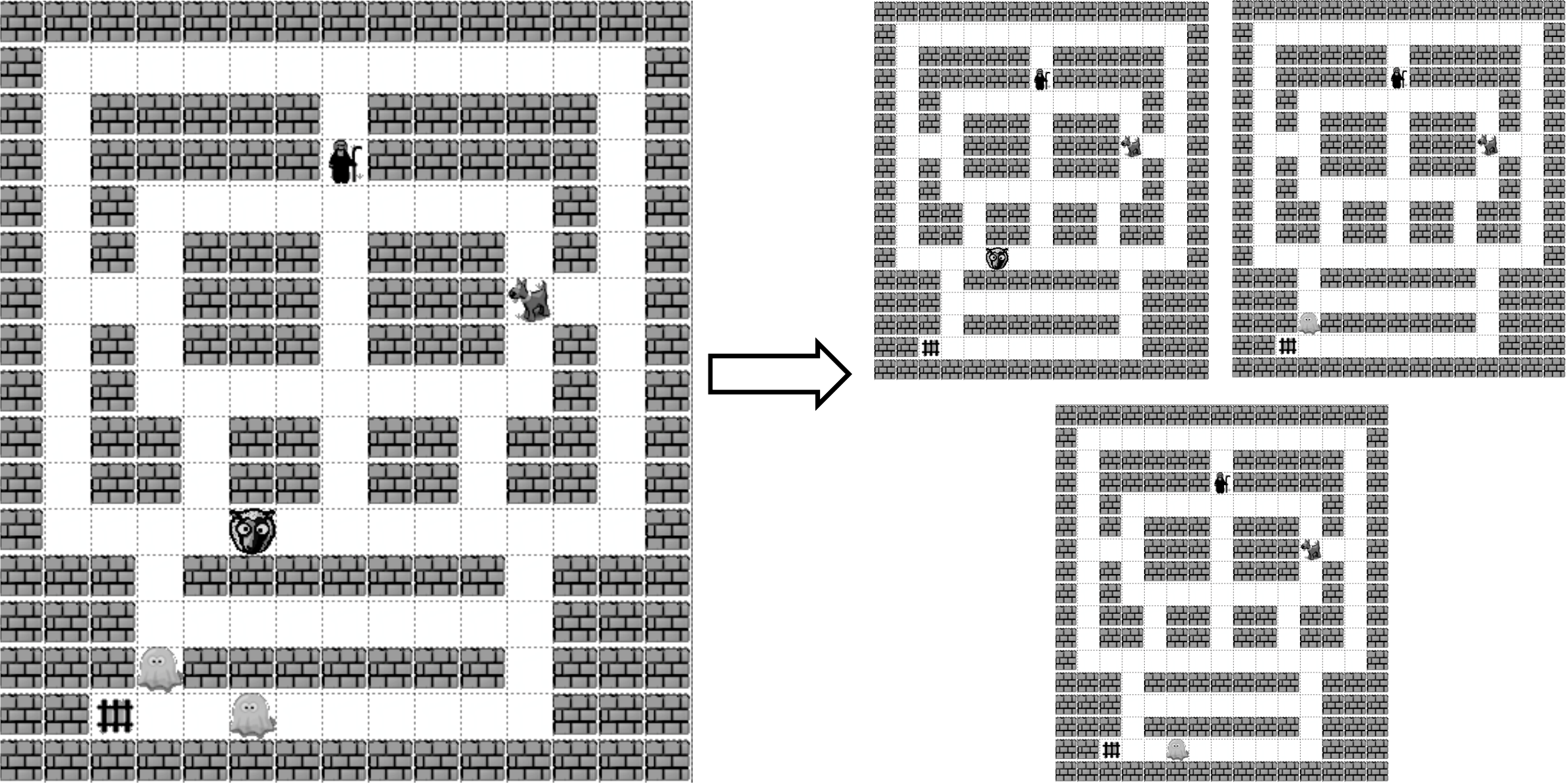} 
\caption{Task decomposition in Sheep Savior.}
\label{fig:sheep-decomp}
\end{figure}

This domain is an extension of the \textit{Collaborative Ghostbuster} game introduced in~\cite{huy2011} as the testbed for their assistance framework for collaborative games. The game features two players (a shepherd and a dog) whose task is to herd a sheep into its pen while avoiding it to be killed by two ghosts in a maze-like environment. All non-player characters (NPCs) run away from the players within a certain distance, otherwise the ghosts chase the sheep and the sheep runs away from ghosts. Since ghosts can only be shot by the Shepherd, the dog's role is strictly to gather the NPCs (Figure~\ref{fig:sheep-decomp}).

Both protagonists have 5 movement actions (no\_move, N, S, E and W) while Shepherd has an additional action to inflict damage on a nearby ghost, hence a total of 30 compound actions. The two players are given rewards for successfully killing ghosts (5 points) or herding sheep into its pen (10 points). If the sheep is killed, the game is terminated with penalty -10. The discount factor in this domain is set to be $ 0.99 $.

\subsection{Choice of heuristic policies}

The game can be seen as having three subtasks, each of which is the task of catching a single ghost or herding a single sheep, as shown in Figure~\ref{fig:sheep-decomp}. Each of these subtasks consists of only two players and one NPC, hence has manageable complexity and can be solved exactly offline using Value Iteration.

A heuristic Q-value can be obtained by taking the average of all individual subworlds', or subtasks', Q-values, as an estimate for one state-action pair's value. Specifically, at state $ s $ the policy, GoalAveraging, yields
\[ n_{GA}(s,a) \leftarrow 1 \]
\[ Q_{GA}(s,a) = \frac{1}{m} \sum_{i=1}^{m} Q_i(s_i,a)\]
in which $ s_i $ is the projection of $ s $ in subtask $ i $, $ m $ is the number of subtasks, i.e. three in this case, and $ Q_i(s_i,a) $ are subtasks' Q-values. The corresponding heuristic policy can be constructed as $ \pi_{GA}(s) = \argmax_a Q_{GA}(s,a)  $.

\textbf{Heuristic quality.} GoalAveraging works well in cases when the sheep is well-separated from ghosts. However, when these creatures are close to each other, the policy's action estimation is no longer valid and could yield deadly results. The under-performance is due to the fact that the heuristic is oblivious to the interactivity between subtasks, in this case, ghost-killing-sheep scenarios. 

\subsection{Setup and results}

The map shown in Figure~\ref{fig:sheep-decomp} is tried 200 times, each of which with a different randomized starting configurations. We compare the means of discounted rewards produced by the following agents: Random, GoalAveraging, UCT, UCT-I, UCT-S, and UCT-Aux. The optimal policy in this domain is not computed due to the prohibitively large state space, i.e., $ 104^5*3^2 \approx 10^{11}$ since each ghost has at most two health points. All UCT variants have a fixed planning depth of $ 300 $. In our setup, one second of planning yields roughly $ 200 $ rollouts on average, so we do not run simulations with higher numbers of rollouts than $ 10000 $ due to time constraint. Moreover, in this game-related domain, the region of interest is in the vicinity of $ 200 $ to $ 500 $ rollouts for practical use.

\begin{figure}[ht]
\centering
\includegraphics[width=230pt]{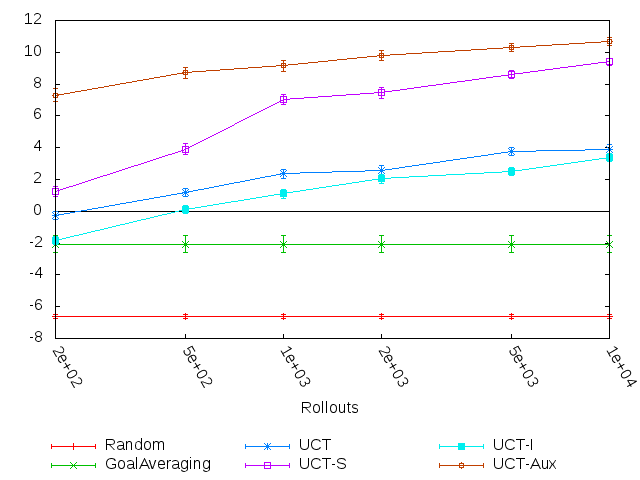}
\caption{Performance comparison of Random, GoalAveraging, UCT, UCT-S, UCT-I, and UCT-Aux when coupled with Goal Averaging.}
\label{fig:sheep-result}
\end{figure}

As shown in Figure~\ref{fig:sheep-result}, UCT-Aux outperforms the other variants, especially early on with small numbers of rollouts. UCT-S takes advantage of GA better than UCT-I, which yields even worse performance than vanilla UCT. Observing the improvement rate of UCT-S we expect it to approach UCT-Aux much sooner than others, although asymptotically all of them will converge to the same optimal value when enough samplings are given and the search tree is sufficiently expanded; the time taken could be prohibitively long though.

\section{Discussions}

Although UCT-Aux shows superior performances in the experiments above, we observe that the chosen heuristic policies share a common property that is crucial for UCT-Aux's success: they show behaviors of ``\textit{make it or break it}". In other words, at most states $ s $, their action value estimate $ Q_{\pi}(s,a) $ is either near-optimal or as low as that of a random policy $ Q_{rand}(s,a) $.

Specifically, in Obstructed Sailing, if following SailTowardsGoal can bring the boat from a starting state to goal position, e.g., when the line of sight connecting source and destination points lies entirely in free space, the resultant course of actions does not deviate much from the optimal action sequences. However when the policy fails to reach the goal, it could be stuck fruitlessly. For instance, Figure~\ref{fig:bad-heu} depicts one such case; once the boat has reached either one of three starred tiles underneath the goal position, unless at least three to five wind directions in a row are E, SailTowardsGoal results in oscillating the boat among these starred tiles. The resultant cost is therefore very far from optimal and could be as low as the cost incurred by random movement. 

\begin{figure}[h]
\centering
\includegraphics[width=200pt]{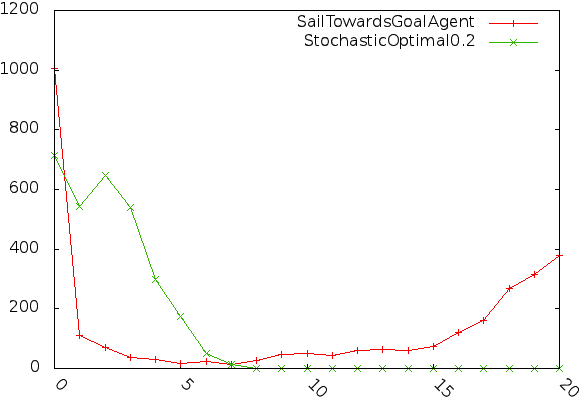}
\caption{Performance histograms of heuristics in Obstructed Sailing. The returned costs of a heuristic are allocated relatively into bins that equally divide the cost difference between Random and Optimal agents; x-axis denotes the bin number and y-axis the frequency.}
\label{fig:heur-sail}
\end{figure}

In contrast, an example for heuristics that are milder in nature is the policy StochasticOptimal.0.2 which issues optimal actions with probability $ 0.2 $ and random actions for the rest. This policy is also suboptimal but almost always yields better estimation than random movement; it is not as ``\textit{extreme}" as SailTowardsGoal. Figure~\ref{fig:heur-sail}, which charts the performance histograms of StochasticOptimal.0.2 alongside with SailTowardsGoal, shows that a majority of runs with SailTowardsGoal yield costs that are either optimal or worse than Random's.

Similarly, GoalAveraging in Sheep Savior is also extreme: By ignoring the danger of Ghosts when around Sheep, it is able to quickly herd the Sheep in the Pen or kill nearby Ghosts (good), or end the game prematurely by forcing the Sheep into Ghosts' zones (bad). We hypothesize that one way to obtain extreme heuristics is by taking near-optimal policies of the relaxed version of the original planning problem, in which aspects of the environment that cause negative effects to the accumulated reward are removed. For instance, SailTowardsGoal is in spirit the same as the optimal policy for maps with no obstacle, while GoalAveraging should work well if the ghosts do not attack sheep.

As UCT-Aux is coupled with heuristic policies with this ``extreme" characteristic, rollouts are centralized at auxiliary arms of states where $ \pi(s) $ is near-optimal, and distributed to ordinary arms otherwise. Consequently, the value estimation falls back to the default random sampling where $ \pi $ produces inaccurate estimates instead of relying entirely on $ \pi $ as does UCT-S. 

\subsection{When does UCT-Aux not work?}

\begin{figure}[h]
\centering
\includegraphics[height=200pt]{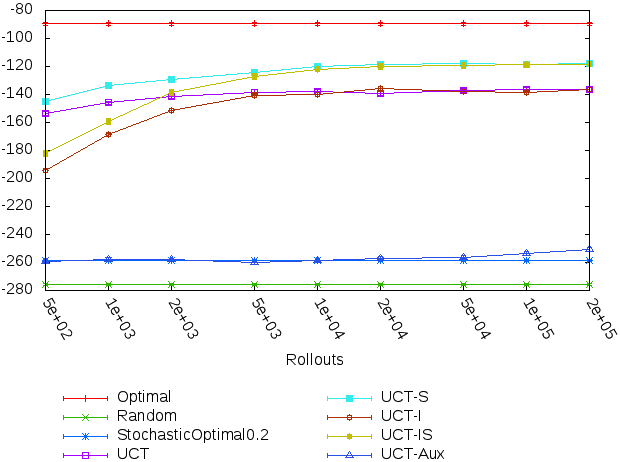} 
\caption{Bad case of UCT-Aux when coupled with StochasticOptimal.0.2.}
\label{fig:sailing-bad}
\end{figure}

Figure~\ref{fig:sailing-bad} charts the worst-case behavior of UCT-Aux when the coupled heuristic's estimate is mostly better than random sampling but much worse than that of the optimal policy, e.g. the heuristic StochasticOptimal.0.2 in Obstructed Sailing.%; in such cases, UCT-S is more appropriate.

%UCT-Aux exhibits worst-case behavior if the coupled heuristic's estimate is mostly better than random sampling but much worse than that of the optimal policy. In such cases, UCT-S is more appropriate, as observed from Figure~\ref{fig:sailing-bad} when UCT variants are coupled with StochasticOptimal.0.2.

The reason behind UCT-Aux's flop is the same as that of UCT, i.e., due to the overly ``optimism" of UCB1, as described in~\cite{coquelin2007}. At each internal node, samplings are directed into suboptimal arms that appear to perform the best so far, exponentially more than the rest (Theorem 1~\cite{kocsis2006}) when convergence has not started. Even though each arm is guaranteed to be sampled an infinite number of times when the number of samplings goes to infinity (Theorem 3~\cite{kocsis2006}), the sub-polynomial rate means only a tiny fraction of samplings are spent on attempting bad-looking arms. As a result, in a specially designed binary tree search case, UCT takes at least $ \Omega(exp(exp(...exp(2)...))) $ samplings before the optimal node is discovered; the term is a composition of $ D-1 $ exponential functions with $ D $ being the number of actions in the optimal sequence.

\begin{table}[hc]
\begin{center}
\begin{tabular} { l | c | c | c | c | c | c | c | c | c }
 Samplings & 500 & 1000 & 2000 & 5000 & 10000 & 20000 & 50000 & 100000 & 200000 \\
\hline \hline
UCT & 268.4 & 555.4 & 1134.6 & 2694.5 & 5395.7 & 11041.7 & 27107.4 & 53235.9 & 107107  \\
UCT-S & 268.9 & 558.6 & 1157.3 & 2733.2 & 5424.7 & 11212.7 & 27851.2 & 54382.2 & 109308 \\
UCT-I & 279.2 & 583 & 1200.7 & 2805 & 5620 & 11519.4 & 28123.8 & 55393.7 & 111418 \\
UCT-IS & 278.6 & 586 & 1209.2 & 2834.8 & 5657.1 & 11704.4 & 28872.5 & 56401.2 & 113682 \\ 
\textbf{UCT-Aux} & \textbf{159.2} & \textbf{256.7} &  \textbf{447.8} & \textbf{889.5} & \textbf{1341.7} & \textbf{1981.3} & \textbf{3117.9} & \textbf{4317.1} & \textbf{5970.11} \\
\end{tabular}
\end{center}
\caption{The average number of tree nodes for UCT variants in Obstructed Sailing when coupled with StochasticOptimal.0.2.}
\label{tab:tree-nodes}
\end{table}
UCT-Aux falls into this situation when coupled with suboptimal policies whose estimates are better than random sampling: At every internal node, it artificially creates an arm that is suboptimal but produces preferable reward sequences when compared to other arms with random sampling. As a result, the auxiliary arms are sampled exponentially more often while not necessarily prescribing a good move.  Table~\ref{tab:tree-nodes} shows some evidence of this behavior: Given the same number of samplings, UCT-Aux constructs a search tree with significantly less nodes than other variants (up to 20 times). That means many samplings have ended up in non-expansive auxiliary arms because they were preferred.

\subsection{Combination of UCT-Aux, UCT-I and UCT-S}

\begin{figure}[h]
\centering
\includegraphics[height=200pt]{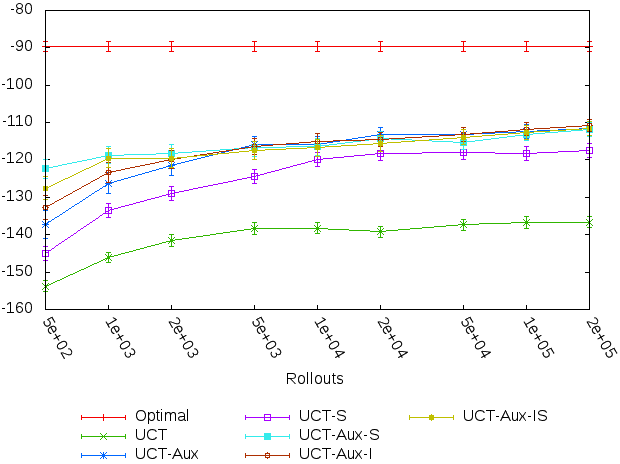} 
\caption{Combination of UCT-Aux with UCT-I/S/IS in Obstructed Sailing.}
\label{fig:sailing-combined}
\end{figure}

UCT-Aux bootstraps UCT in an orthogonal manner to UCT-I and UCT-S, thus allowing combination with these common techniques for further performance boost when many heuristics are available. Figure~\ref{fig:sailing-combined} charts the performance of such combinations in Obstructed Sailing. UCT-Aux variants use SailTowardsGoal at the auxiliary arms while UCT-I/S variants use StochasticOptimal.0.2 at the ordinary arms. UCT-Aux-S outperforms both UCT-Aux and UCT-S at earlier stage, and matches the better performer among the two, i.e. UCT-Aux, in a long run.

\section{Conclusion}

In this work, we have introduced a novel yet simple technique to bootstrap UCT with an imperfect heuristic policy in a popular non-adversarial domain, i.e., planning in large-state space MDPs. It is shown to be able to leverage on the well-performing region while avoiding the bad regions of the policy, empirically outperforming other state-of-the-art bootstrapping methods when coupled with the right policy, i.e, the ``extreme" kind. Our conclusion is that if such property is known before hand about a certain heuristic, UCT-Aux can be expected to give a real boost over the original UCT, especially in cases with scarce computational resource; otherwise, it would be safer to employ the currently prevalent methods of bootstrapping. As such, a different mentality can be employed when designing heuristics specifically for UCT-Aux: instead of safe heuristics that try to avoid as many flaws as possible, the designer should go for greedier and ``riskier" ones. Lastly, since UCT-Aux is orthogonal to other commonly known enhancements, it is a flexible tool that can be combined with others, facilitating more options when incorporating domain knowledge into the vanilla MCTS algorithm. In the future, we plan to examine how to adapt the method to adversarial domains.

\section{Acknowledgments}

This research is partially supported by a GAMBIT grant "Tools for Creating Intelligent Game Agents", no. R-252-000-398-490 from the Media Development Authority and an Academic Research Grant no. T1 251RES1005 from the Ministry of Education in Singapore.% The authors would like to thank Ye Nan (NUS) for various valuable discussions while developing and evaluating the proposed method, and Tomi Silander (NUS) for his insightful comments on how to improve the paper's presentation.%, and the reviewers for their constructive criticism on the paper.

\bibliographystyle{splncs}
\bibliography{truonghuy}

\end{document}